%% file: Virtual_ID_at_alibaba.tex
\documentclass[sigconf]{acmart}

\PassOptionsToPackage{pass}{geometry}
\usepackage{booktabs} 
\usepackage{multirow}
\usepackage{algorithm}
\usepackage{algorithmic}
\usepackage{amsmath}
\usepackage{xcolor}
\usepackage{color}

\copyrightyear{2019}
\acmYear{2019}
\setcopyright{acmcopyright}
\acmConference[CIKM '19]{The 28th ACM International Conference on Information and Knowledge Management}{November 3--7, 2019}{Beijing, China}
\acmBooktitle{The 28th ACM International Conference on Information and Knowledge Management (CIKM '19), November 3--7, 2019, Beijing, China}
\acmPrice{15.00}
\acmDOI{10.1145/3357384.3357800}
\acmISBN{978-1-4503-6976-3/19/11}

\begin{document}
\title{Virtual ID Discovery from E-commerce Media at Alibaba}
\subtitle{Exploiting Richness of User Click Behavior for Visual Search Relevance}




\author{Yanhao Zhang, Pan Pan, Yun Zheng, Kang Zhao, Jianmin Wu, Yinghui Xu, Rong Jin}
\affiliation{%
  \institution{Machine Intelligence Technology Lab, Alibaba Group}
}
\email{yanhao.zyh, panpan.pp, zhengyun.zy, zhaokang.zk, jianmin.jm, renji.xyh, jinrong.jr@alibaba-inc.com}




\renewcommand{\shortauthors}{Zhang et al.}
\renewcommand{\algorithmicrequire}{\textbf{Input:}}
\renewcommand{\algorithmicensure}{\textbf{Output:}}

\begin{abstract}
Visual search plays an essential role for E-commerce. To meet the search demands of users and promote shopping experience at Alibaba, visual search relevance of real-shot images is becoming the bottleneck. Traditional visual search paradigm is usually based upon supervised learning with labeled data. However, large-scale categorical labels are required with expensive human annotations, which limits its applicability and also usually fails in distinguishing the real-shot images.

In this paper, we propose to discover Virtual ID from user click behavior to improve visual search relevance at Alibaba. As a totally click-data driven approach, we collect various types of click data for training deep networks without any human annotations at all. In particular, Virtual ID are learned as classification supervision with co-click embedding, which explores image relationship from user co-click behaviors to guide category prediction and feature learning. Concretely, we deploy Virtual ID Category Network by integrating first-clicks and switch-clicks as regularizer. Incorporating triplets and list constraints, Virtual ID Feature Network is trained in a joint classification and ranking manner. Benefiting from exploration of user click data, our networks are more effective to encode richer supervision and better distinguish real-shot images in terms of category and feature. To validate our method for visual search relevance, we conduct an extensive set of offline and online experiments on the collected real-shot images. We consistently achieve better experimental results across all components, compared with alternative and state-of-the-art methods.

\end{abstract}

%
%


\maketitle
\input{introduction}

\input{mainbody}

\input{experiment}

\bibliographystyle{ACM-Reference-Format}
\bibliography{visual-ID-bib}
\end{document}

%% file: introduction.tex
\section{Introduction}

Due to the explosive growth of online photos in search engines and social media, Virtual ID discovery aims at learning rich and emerging labels for deep networks instead of human annotations. Traditionally, data-driven classification and feature learning paradigms require large datasets with supervised categorical labels to train properly. However, collecting such large-scale dataset is challenging, for instance, 1) Domain knowledge requirement: domain experts are needed to label images for specific domains, which may be hard to find via crowdsourcing. 2) Lack of large-scale labeled datasets: huge expense of collecting data limits the applicability of the supervised methods. Besides, the definitions and labeling of attributes are very troubling, which consumes a lot of manpower and financial resources. 3) Cross-domain feature matching: traditional collection and training with human labeled data can't solve the feature matching problem between online product images and offline users' real-shot images.

\begin{figure}[t]
\begin{center}
\includegraphics[width=1.0\linewidth]{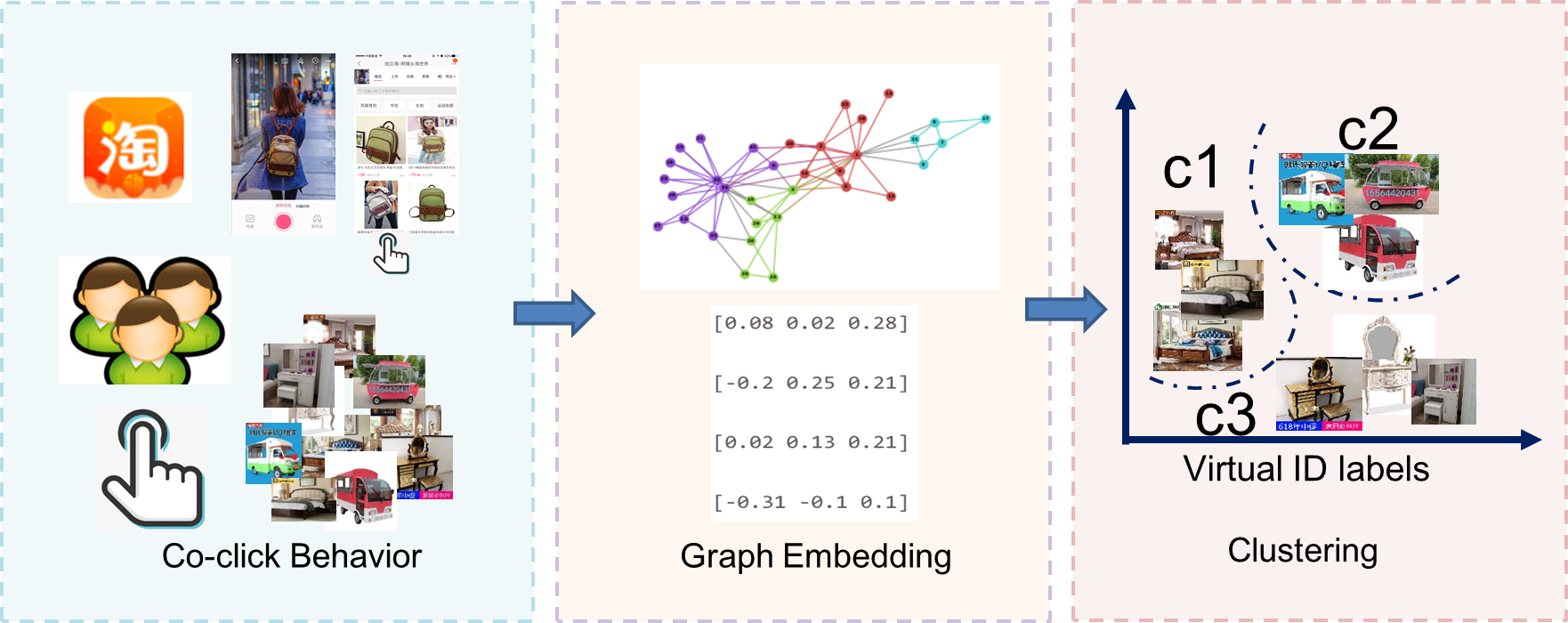}
\caption{The illustration of Virtual ID discovery at Alibaba: In this paper, we answer that how to discover Virtual ID from user click behavior and how to employ ``crowdsourcing'' annotations for improving the relevance in Pailitao\protect\footnotemark, Alibaba's visual search application~\cite{zhang2018visual}.}
\label{fig:1_intro}
\end{center}
\end{figure}
\footnotetext{http://www.pailitao.com}

Fortunately, user behavior data contains large collective intelligence. Pinterest~\cite{zhai2017visual} collects content data from corresponding activities of users, such as likes, comments, or shares to enhance new products.  In Microsoft, click-through data as relevance feedback are also employed for text search and video tagging to identify search intention~\cite{pan2014click,yao2013annotation}. Subsequently, exploiting the user click behavior in E-commercial systems is imperative, due to the obvious advantages, 1) massive data to be collected, 2) natural scenarios to label various images. 3) latent relationship of images to be mined. User click behavior data is aggregated across many users, which may recover interesting properties of the content. Considering the real-world visual search systems Microsoft~\cite{hu2018web}, Ebay~\cite{kdd_YangKBSWKP17}, Pinterest~\cite{kdd_JingLKZXDT15} release their research methods to describe the algorithms and deployment, there are few works describing exploration of click behavior for commercial image search applications in detail.
\begin{figure*}[t]
\begin{center}
\includegraphics[width=1.0\linewidth]{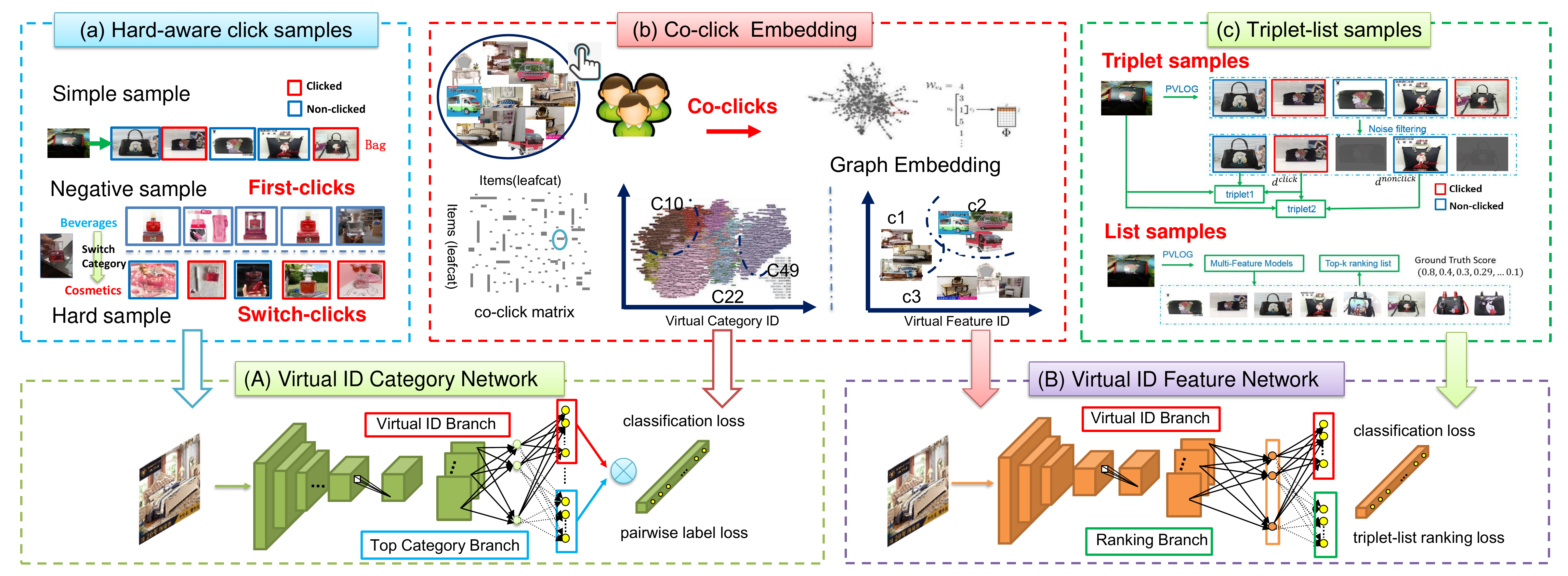}
\caption{Overview of the exploration of user click behavior and Virtual ID Networks architecture.}
\label{fig:3_framework}
\end{center}
\end{figure*}

At Alibaba, we also face the issue on how to deal with massive data to improve visual search relevance for ``Pailitao''~\cite{zhang2018visual}. As the world's biggest visual search system, ``Pailitao'' triggered great attention and widespread recognition in industry, and has experienced swift growth with over 20 million Daily Active User(DAU) in 2018. Millions of users are taking photos, browsing, clicking, and purchasing etc, leaving a huge amount of behavior data as implicit feedback everyday. The essential objective of visual search at Alibaba aims at finding the identical product, especially in case of real-shot scenarios. Therefore, the demand is clear to optimize the search results, meaning that accurate matching the identical products for real-shot images is the top priority. To achieve this, we resort to click-data driven approach for visual search relevance by leveraging the collective human intelligence that is underlying user click behavior. As we all know, manually labeling data is expensive while the click logs are not that expensive to obtain.

In this paper, we would like to share the click-data driven developments as shown in Figure~\ref{fig:1_intro}, explicitly addressing the existing challenges at Alibaba. \textbf{What is Virtual ID discovery? the main idea is to learn supervised labels from co-click behavior for both category prediction and feature learning collaboratively}. In particular, we explore the co-occurrence similarity of images by clustering the graph embedding of co-clicks behavior. As the major departure from the existing paradigms, we do not rely on any human annotated labels but employ Virtual ID as classification labels for training the deep networks. As the observation, most co-click data are long-tailed distributed and sometimes are erroneous over the images. Additional click behaviors are introduced to reveal potential supervision from user interactions and can be exploited to improve the performance.

Addressing the challenges above, we present Alibaba's click-data driven approach for visual search relevance in detail. We illustrate the algorithm of Virtual ID discovery and take a step further to feed various click data into training the deep networks. Specifically, we describe the details of how we leverage Virtual ID Category and Feature Networks in terms of user clicks data. We conduct extensive experiments on real-shot dataset to evaluate the effectiveness of the proposed visual search modules.

\section{Related work}
We briefly group the related work into two aspects: data-driven feature learning, and search by mining click data.

Data-driven features are shown to effectively encode both appearance and semantics information, which indicates superiority on many recognition benchmarks~\cite{cvpr_HeZRS16,cvpr_SzegedyLJSRAEVR15,krizhevsky2012imagenet}. However, lots of labeled images are required for deep learning based models to train properly. Unsupervised feature learning approaches~\cite{yuan2014unified,fang2015collaborative,zhang2017learning} are applied, which achieve significant promise in terms of overcoming the human annotation limitation. Our work is most related to~\cite{yuan2014unified,fang2015collaborative}, which mine latent factors from social media as CNN supervision. Different from~\cite{fang2015collaborative}, our work focuses on how to learn Virtual ID and integrate various types of user click data for training. In~\cite{zhang2017learning}, the resultant image-word embedding spaces are learned from the social multimedia content. Our Virtual ID supervision differs from ~\cite{zhang2017learning} and is acquired by adopting the graph embedding of co-click data, which encodes implicit feedback for image recognition and retrieval task.

User behavior data has been studied and analyzed widely with various web mining techniques to improve the efficacy and usability of search system~\cite{chu2018deep,pan2014click}. Befferman et al.~\cite{beeferman2000agglomerative} suggested an agglomerative clustering to identify related queries by the use of the click-through data. Li et al.~\cite{li2008learning} presented the approach using click graphs to improve query intent classifiers. Several approaches have tried to model the representation of queries or documents on the click-through bipartite graph~\cite{pan2014click,yao2013annotation}. Unlike most of the above approaches that focus on leveraging both click data and features only from the textual view, our work aims to discover relations from co-click graph combined with abundant click behavior to train the deep networks for image search purpose.

Despite success of above works, challenges and issues still exist on how to apply different click behaviors from specific scenarios for the goal of searching the most relevant items. At Alibaba, it is non-trivial to deal with billions of data and perform satisfying performance for users' intention. With these realistic challenges in mind, we conclude our contributions as following:

\textbf{(a)} We present a click-data driven approach to discover Virtual ID for deep networks training, which achieves better performance and scalability for virtual search relevance. Image relationship underlying the co-clicks are mined to guide category prediction and feature learning. Specifically, we learn graph embedding from co-clicks and generate Virtual ID by clustering their corresponding co-click embeddings. To the best of our knowledge, this is the first comprehensive study of employing user click behavior for visual search problem without any human annotations.

\textbf{(b)} By collecting the various types of user click behavior, we deploy Virtual ID Category and Feature Networks to improve search relevance collaboratively. Integrating with the first-clicks and switch-clicks behaviors, we apply Virtual ID Category Network with Virtual ID as classification labels to characterize the confusable categories. Similarly, we propose Virtual ID Feature Network to simultaneously learn discriminative feature from Virtual ID supervision and preserve ranking from triplet and list samples.

\textbf{(c)} Extensive experiments are thoroughly conducted in offline and online setting, which demonstrate the effectiveness of Virtual ID Networks. As deployed on visual search for millions of users, online A/B testing shows visual search results returned by our models meet users' preference better and produce greater business value.

\begin{figure}[t]
\begin{center}
\includegraphics[width=1.0\linewidth]{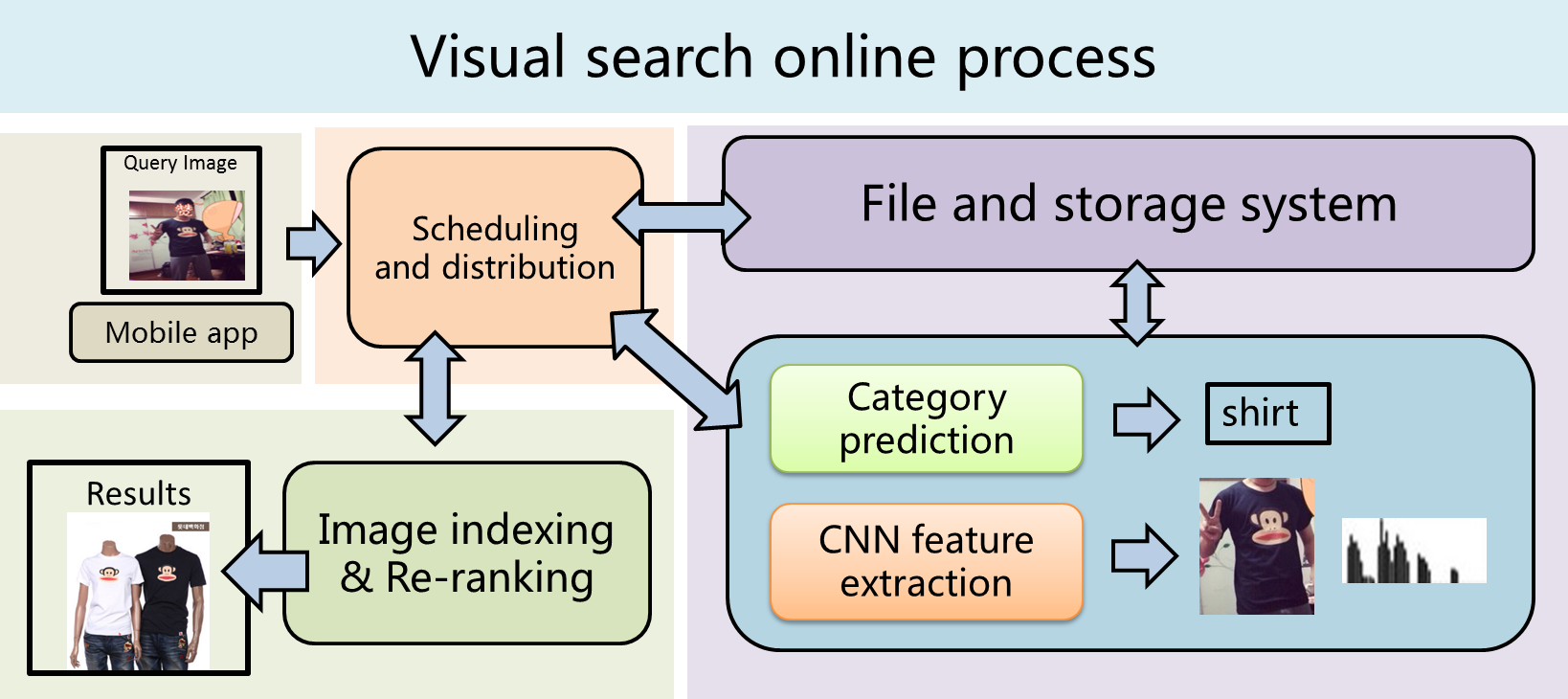}
\caption{Visual search online process at Alibaba.}
\label{fig:2_onlineprocess}
\end{center}
\end{figure}






%% file: mainbody.tex
\section{Visual search relevance}
Visual search aims at retrieving images by visual features for relevant image list. As shown in Figure~\ref{fig:2_onlineprocess}, by continuous polishing of product technology, we settle down stable and scalable online process~\cite{zhang2018visual}. When users uploaded query images, two schemes are closely related to visual search relevance, referred as \textbf{category prediction} and \textbf{feature extraction}. The reasons can be attributed to that 1) we certainly can't find the identical products once falling into the wrong predicted category. 2) Capability of CNN feature is no doubt the key step to obtain the most relevant results.


Click behavior data is not just bridging user intention gap for image search, but also serve as reliable and implicit feedback for training deep networks. Therefore, Virtual ID is able to reveal latent relationship by exploring the co-occurrences of the users' clicks. Figure~\ref{fig:3_framework} illustrates the unified structure of Virtual ID discovery in Pailitao. In Figure~\ref{fig:3_framework}(A-B), Virtual ID are fed into Virtual ID Category and Feature Networks by jointly integrating various types of click data from Figure~\ref{fig:3_framework}(a-c). Specifically, a bipartite graph between the users and clicked items is constructed based on the PVLOG from image search engine. Moreover, leaf-category and image spaces are formed by co-click graphs respectively. The link between every two nodes in each space represents the leaf-category or image similarity. The spirit of Virtual ID is to learn pseudo labels by means of clustering their corresponding latent embeddings, while preserving the inherent structure of co-click graph. Taking advantages of Virtual ID, deep networks could be trained based on classification labels in the traditional supervised way. In the following, we will cover details of Virtual ID algorithm.

\subsection{Discovering Virtual ID from Co-clicks}
\label{sec:virualID}
Let $\mathbf{G}=(\mathbf{V}, \mathbf{E})$ denotes the co-click graph constructed from user click-through bipartite graph $\mathbf{U}$. $\mathbf{V}$ is the set of vertices, which consists of the clicked items in the returned list from the users. $\mathbf{E}$ is the set of edges between item vertices. An edge between two images on the returned list is established when the users who issued the query clicked both images.

Our goal is to learn the clusters of graph embedding that can well capture semantic relationship of the vertices. As the click data is usually long-tailed distributed and noisy, spectral clustering on graph is not capable of achieving adaptive clusters. By exploring diverse neighborhoods of vertices, we apply DeepWalk~\cite{perozzi2014deepwalk} from truncated random walks to learn representations of vertices in co-click graph. The distribution of nodes appearing in short random walks appears similar to the distribution of words in natural language. Skip-Gram model as the word2vec model, is adopted by DeepWalk to learn the representations of nodes. Specifically, for each walk sequence $s={v_1, v_2,\ldots,v_s}$, following Skip-Gram, DeepWalk aims to maximize the probability of the context of node $v_i$ in this walk sequence as follows:
\begin{eqnarray}\label{eqn:triplet_loss1}
\max_{\Theta}Pr(\{v_{i-w},\ldots, v_{i+w}\}\setminus v_i|\Theta(v_i))\\\nonumber
=\prod^{i+w}_{j=i-w,j\neq i}Pr(v_j|\Theta(v_i))
\end{eqnarray}
where $w$ is the window size, $\Theta(v_i)$ denotes the current representation of $v_i$ and $\{v_{i-w},\ldots, v_{i+w}\}\setminus v_i$ is the local neighbourhood nodes of $v_i$. hierarchical soft-max is used to efficiently infer the embeddings. Finally, we cluster the graph embedding $\Theta(\mathbf{V})$ into $K$ clusters as Virtual ID $\{C_1,\ldots,C_K\}$.

\subsection{Virtual ID Category Network}
\label{sec:catpre}
In this part, we explain how to train Virtual ID Category Network from the user click data. Considering both visual and semantic similarity, query image is firstly recognized into 14 top categories, such as shoes, dress, bags etc, which covers all the leaf-categories. As hierarchy system of leaf-categories, the network contains two branches in terms of supervised data for Virtual ID and top category in Figure~\ref{fig:3_framework}(A).
\begin{figure}[t]
\begin{center}
\includegraphics[width=1.0\linewidth]{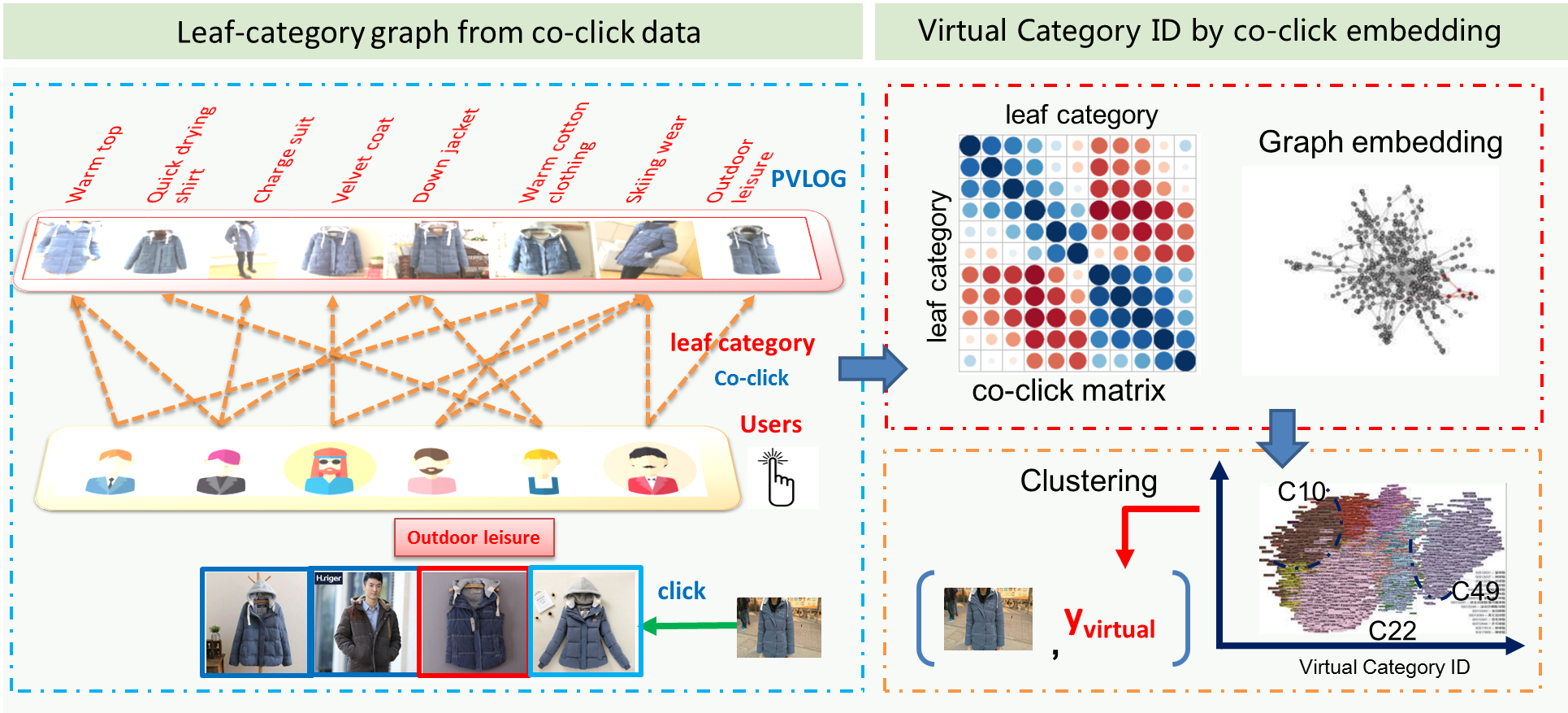}
\caption{Co-clicks mining for Virtual Category ID.}
\label{fig:4_coclick_cate_graph}
\end{center}
\end{figure}

\textbf{Leaf-category graph by co-clicks.}
To distinguish the confusable categories more accurate, we construct the co-click graph by using leaf-categories attached with clicked image, as shown in Figure~\ref{fig:4_coclick_cate_graph}. Some top categories contain thousands of leaf-categories, which are easily confused with each other. 1) For similar leaf-categories, we attempt to merge the fined-grained leaf-categories into virtual category node. 2) For the coarser top categories, we split them into multiple virtual category nodes. Following Section~\ref{sec:virualID}, we collect clicked image with corresponding Virtual Category ID to supervise the Virtual ID branch in the network. For the confusable top categories, we make a decision by mapping the predicted Virtual ID back to the corresponding top category. In Figure~\ref{fig:4_coclick_cate_graph}, the network is trained using each query image $\mathbf{q}$ associated with the label $\mathbf{y}_\text{virtual}$. We deploy ResNet50~\cite{cvpr_HeZRS16} network for trade-off between high accuracy and low latency. Standard softmax-loss is employed for Virtual ID branch in terms of classification task.
\begin{eqnarray}
\label{eqn:triplet_loss}
\mathcal{L}_\text{virtual}(\mathbf{q},\mathbf{y}_\text{virtual})=-\log(e^{\mathbf{q}_{\mathbf{y}_\text{virtual}}}/ \sum_j^N e^{\mathbf{q}_j})
\end{eqnarray}

\begin{figure}[t]
\begin{center}
\includegraphics[width=1.0\linewidth]{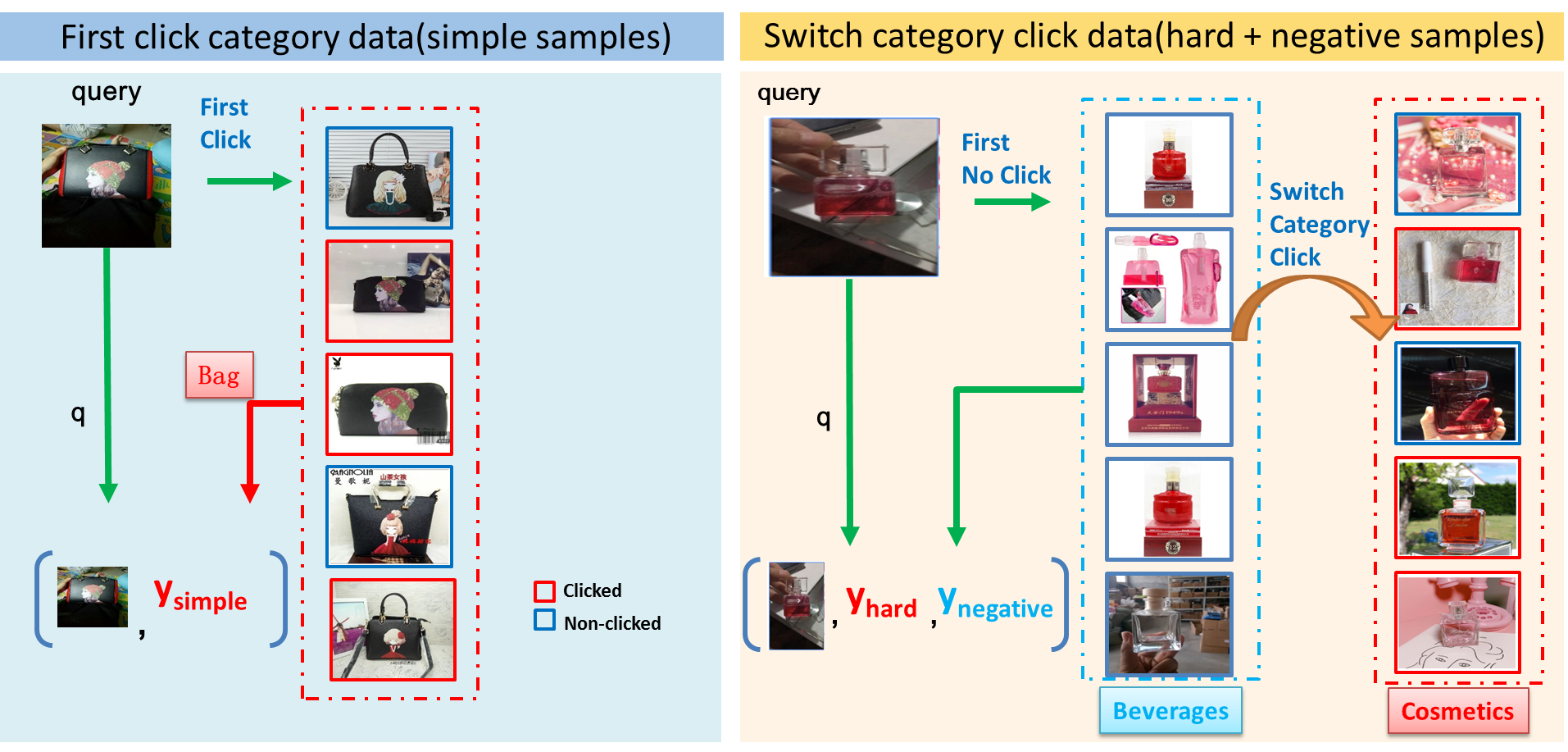}
\caption{First-clicks and switch-clicks for top category branch.}
\label{fig:5_first_switch_clicks}
\end{center}
\end{figure}

\textbf{First-clicks as simple samples.}
To regularize the learned Virtual ID branch, we further make use of the top category clicks accompanying each query image $\mathbf{q}$ and feed them into the top category branch. As shown in Figure~\ref{fig:5_first_switch_clicks}, we can easily collect a large amount of first-click data and form simple samples containing the query image $\mathbf{q}$ with the top category label $\mathbf{y}_\text{simple}$.

\textbf{Switch-clicks as hard and negative samples.}
Notice that most of the users' effective behavior such as switching clicks can correct the errors of the previous category predictions. Therefore, we mine the PVLOG of the switching category behavior in Figure~\ref{fig:5_first_switch_clicks} and collect the previous non-click category as the negative sample label $\mathbf{y}_\text{neg}$. After switching, the clicked category can be treated as hard sample label $\mathbf{y}_\text{hard}$. At this point, pairwise label loss can be formed for training,

\begin{eqnarray}\label{eqn:triplet_loss1}
\mathcal{L}_\text{pair}(\mathbf{q}, \mathbf{y}_\text{neg},\mathbf{y}_\text{hard})=\max (0,(1-\mathbf{q}_{\mathbf{y}_\text{neg}} + \mathbf{q}_{\mathbf{y}_\text{hard}})^2)
\end{eqnarray}
To encourage diversity of the training samples, hard samples play a more important role in contributing to classification performance. Click samples $\mathbf{y}_\text{click}$ are collected containing simple $\mathbf{y}_\text{simple}$ and hard $\mathbf{y}_\text{hard}$ samples and further added by an indicator $\mathcal{H}$ to comprise hard-aware samples. The top category loss is then,
\begin{eqnarray}\label{eqn:triplet_loss1}
\mathcal{L}_\text{hard-aware}(\mathbf{q},\mathbf{y}_\text{click})=\mathcal{H}(-\log(e^{\mathbf{q}_{\mathbf{y}_\text{click}}}/ \sum_j^N e^{\mathbf{q}_j}))
\end{eqnarray}
The final training loss therefore becomes:
\begin{eqnarray}
\label{eqn:loss_cat}
\mathcal{L}_\text{Cat}=\mathcal{L}_\text{virtual}+\alpha\mathcal{L}_\text{hard-aware}+\beta\mathcal{L}_\text{pair}
\end{eqnarray}
where $\alpha$, $\beta$ are scalar parameters. We trained the two branches together according to $\mathcal{L}_\text{Cat}$. Benefiting from the these types of click behavior, we ensemble the results of two branches and make the final prediction. Overall, the result brings over 9\% absolute improvement on Top-1 accuracy for category prediction of real-shot images.

\subsection{Virtual ID Feature Network}
\label{sec:joinmodel}
In this section, we will introduce the feature learning network based on user click behavior, which is formulated as a joint classification and ranking problem. Despicted in Figure~\ref{fig:3_framework}(B),  a classification loss with Virtual feature ID is employed on the Virtual ID branch, while a ranking loss with triplet-list samples is incorporated on the ranking branch. To maximum extent, we take advantage of the user click data to encode similarity relationship and ranking orders for discriminative feature training. As a result, we construct Virtual ID Feature Network that is able to jointly learn classification and ranking from clicked data without any annotations.
\begin{figure}[t]
\begin{center}
\includegraphics[width=1.0\linewidth]{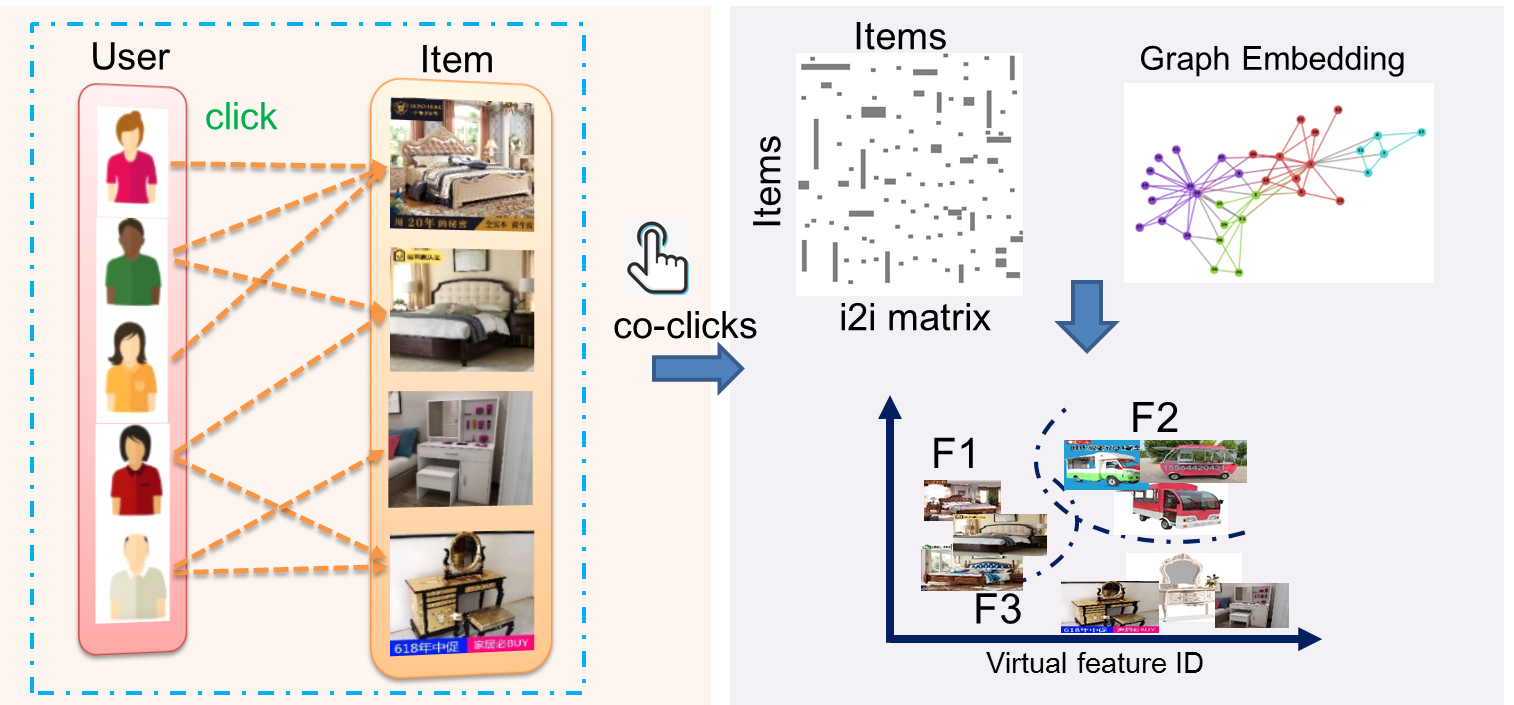}
\caption{Co-clicks mining for Virtual Feature ID.}
\label{fig:5_coclick_item_graph}
\end{center}
\end{figure}

\textbf{Item graph by co-clicks.}
Inspired by the CNN on large-scale ImageNet~\cite{ILSVRCarxiv14} being able to learn quite generic visual features, we apply the co-clicks to create Virtual Feature ID as pseudo classes for query images and then learn high-level features in a supervised way. As shown in Figure~\ref{fig:5_coclick_item_graph}, we construct the co-click graph using image items rather than leaf-categories as nodes and follow Section~\ref{sec:virualID} to form training samples containing query image $\mathbf{q}$ and Virtual Feature ID $\mathbf{y}_\text{virtual}$. Accordingly, we assign the same ID for more similar images according to item-level co-click embedding. Even though item graph contains noise, the learned Virtual Feature ID can greatly reflect semantic and visual relationship between our content items. In order to increase the capability of the learned visual feature, it is necessary to increase the number $K$ of Virtual ID $\{C_1,\ldots,C_K\}$ to partition feature space of images to more fine-grained clusters. By reflecting Virtual Feature ID based on the cluster index of co-click embedding, we employ ResNet50 to learn such K-way Virtual ID branch in Feature Network.

\textbf{Triplets and list constraints for ranking.}
While the classification loss by itself should be sufficient to learn discriminative feature, we note that in practice it is critical to complement with ranking loss simultaneously. Although clicked list data source can be used in different forms to learn visual features, we resort to a more emerging supervised way by creating triplets and list samples(Figure~\ref{fig:3_framework}(c)) from the clicked items in the returned list.

\begin{figure}[t]
\begin{center}
\includegraphics[width=1.0\linewidth]{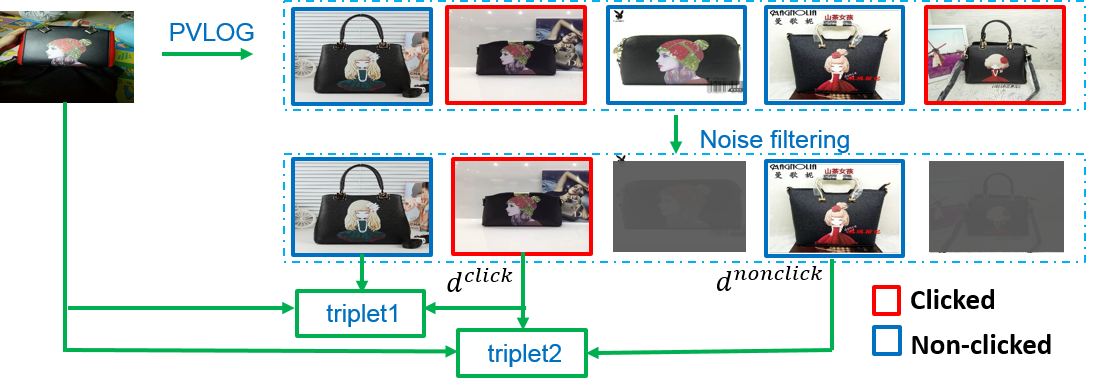}
\caption{Triplets sample mining from PVLOG.}
\label{fig:6_tripletclick}
\end{center}
\end{figure}

Specifically, we can easily obtain a set of triplets $\mathcal{T}$, where each tuple $(\mathbf{q}, \mathbf{q}^+, \mathbf{q}^-)$ consists of an input query $\mathbf{q}$, a similar image $\mathbf{q}^+$ and a dissimilar image $\mathbf{q}^-$. As shown in Figure~\ref{fig:6_tripletclick}, we assume that the clicked images $d^\text{click}$ are more likely to be identical than the non-click images $d^\text{nonclick}$. To preserve the similarity relations in the triplets, we aim to learn the CNN embedding $\mathbf{f}(.)$ which makes the positive image $\mathbf{f}(\mathbf{q}^+)$ more similar to $\mathbf{f}(\mathbf{q})$ than the negative image $\mathbf{f}(\mathbf{q}^-)$. Therefore, triplet loss is used as,
\begin{eqnarray}\label{eqn:triplet_loss1}
&&\mathcal{L}_\text{triplet}(\mathbf{q}, \mathbf{q}^+, \mathbf{q}^-)\\\nonumber
&=&\max(0,||\mathbf{f}(\mathbf{\mathbf{q}})-\mathbf{f}(\mathbf{q}^+)||_2-||(\mathbf{f}(\mathbf{q})-\mathbf{f}(\mathbf{q}^-)||_2+1)
\end{eqnarray}


For triplet ranking loss, hard sample mining is a critical criterion to guarantee its performance in efficient similarity search. Given certain returned list from user clicks scenarios, we observed the users click the identical product images $d^\text{click}$, which can be regarded as the query's positive images. The identical images lie in non-clicked images $d^\text{nonclick}$ would badly affect its performance. By removing false negatives, we adopt a multi-feature fusion approach to filter the non-clicked identical images according threshold $\gamma$ as, $\mathbf{q}^-\in\{{d^\text{nonclick}|\min[\text{D}(d^\text{nonclick},\mathbf{q}), \text{D}(d^\text{nonclick},d^\text{click})] \geq \gamma}\}$.
To ensure noisy negatives to be found more accurately, we combine the local feature, previous version features and pre-trained ImageNet~\cite{ILSVRCarxiv14} feature to compute the integrated distance $\text{D}(.)$. More accurate positive images are obtained following similar process, $\mathbf{q}^+\in\{d^\text{click}|\text{D}(d^\text{click},\mathbf{q})\leq \varepsilon\}$.

\begin{figure}[t]
\begin{center}
\includegraphics[width=1.0\linewidth]{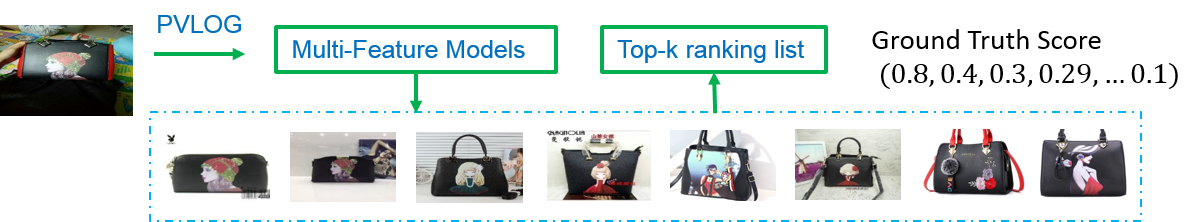}
\caption{List sample mining from PVLOG.}
\label{fig:7_listclick}
\end{center}
\end{figure}

To further inherit high ranking performance from multi-feature fusion, we minimize a listwise ranking loss with ground truth ranking in a mini-batch. As the teacher models, multi-feature models are able to capture more patterns from data and thus has a strong performance. To obtain the ground truth in Figure~\ref{fig:7_listclick}, we compute the predicted relevance scores of the teacher models for returned images and get the top-N ranking permutation $\{\pi_1, \ldots, \pi_N\}$. The probability of a ranking list is computed as,
\begin{eqnarray}\label{eqn:triplet_loss}
P(\pi|X)=\prod^N_{i=1}\frac{\mathcal{W}_i\exp[\text{D}(x_{\pi(i)},\mathbf{q})]}{\sum^N_{k=i}{\exp[\text{D}(x_{\pi(k)},\mathbf{q})]}}
\end{eqnarray}
where the assigned relevance scores $\mathbf{d}$ are computed by $\text{D}(.)$ on the learned feature vector $x_\pi(i)$, $\mathcal{W}_i$ denotes the importance of the position. With the cross entropy as the metric, listwise ranking loss is improved as,
\begin{eqnarray}\label{eqn:triplet_loss}
\mathcal{L}_\text{listwise}(\mathbf{q},\pi,x) = -\sum_{\pi\in\mathcal{P}} P(\pi|\mathbf{d})\log P(\pi|x)
\end{eqnarray}

This list-based constraint makes better use of non-clicked false negatives, which preserves relative position within the rank. It brings an improvement beyond the triplets, as well as keeping the learning stable by considering more ranked pairs, since only using the hard triplets can in practice lead to bad local minima in training.

\textbf{Joint classification and ranking.}
We deploy the two branches of Virtual ID Feature Network with balance scalar $\lambda$ by combining classification loss as well as triplet-list loss:
 \begin{eqnarray}\label{eqn:triplet_loss}
\mathcal{L}_\text{Fea} = \mathcal{L}_\text{virtual}+\lambda(\mathcal{L}_\text{triplet}+\mathcal{L}_\text{listwise})
\end{eqnarray}
By feeding $(\mathbf{q}, \mathbf{q}^+, \mathbf{q}^-)$ as the triplets and $(\mathbf{q},\pi)$ as the list simultaneously, we maximize the positive and negative characteristics in triplets and preserve the informative ranking.

Finally, we deploy Virtual ID Category Network to predict certain category, and extract the 512-dim CNN feature of FC layer with Virtual ID Feature Network for online process. In practice, we found that user co-click behavior data encodes both visual information and crowd semantic properties. Discrete Virtual ID, which is essentially a vector quantization coding, could be robust to outliers using softmax-loss. Further, associating with various click data in networks makes great improvement for the search relevance. In our experiments, we report that learning from Virtual ID can produce satisfactory category prediction and visual features.

%% file: experiment.tex
\section{Experiment and Analysis}
In this section, we conduct extensive experiments to evaluate the performance of category prediction and feature learning paradigms in our system, which follow the same protocol in~\cite{zhang2018visual}. We take the ResNet50 model~\cite{cvpr_HeZRS16} as the backbone for both Virtual ID Networks.

\textbf{Dataset.} To conduct evaluation for both relevance related components in visual search, we collected 200 thousand real-shot images along with top categorical labels and the identical item labels of retrieved results as ground truth. Our evaluation set covers various real-shot images in 14 top categories as listed in Table~\ref{tb:CateResult}. We illustrate comprehensive evaluation result of the components in the unified architecture with various evaluation metrics.

\begin{figure}[t]
\begin{center}
\includegraphics[width=1.0\linewidth]{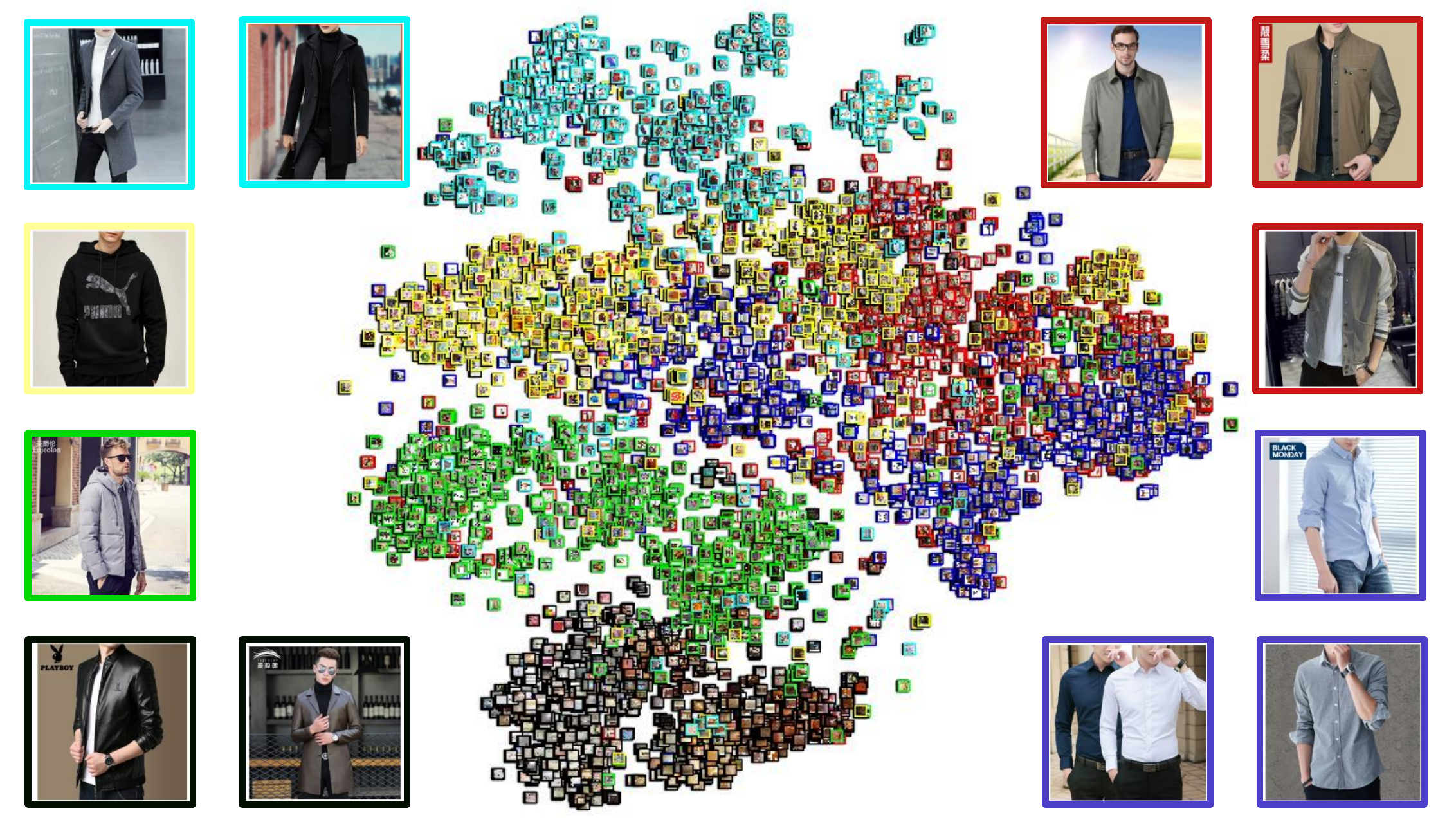}
\caption{Visualization~\cite{van2008visualizing} of 512-dim embedding for 6 leaf-categories in ``shirt'' category (best viewed in color).}
\label{fig:tsne}
\end{center}
\end{figure}
\subsection{Evaluation of Category Prediction}
We conduct experiments to evaluate the performance of category prediction from Virtual ID Category Network, containing Virtual ID and top category branches. In Table~\ref{tb:CateResult}, we list the results of the separated approaches in terms of Precision@1 of top category and conclude the average accuracy. As we can see, our Virtual ID Category Network results in better category prediction than either individual branch. Our Virtual ID branch achieves average Top-1 accuracy 76.48\%, which already surpasses the online model~\cite{zhang2018visual} with accuracy 72.88\%. Besides, top category branch achieves slightly higher results than Virtual ID branch in some categories, i.e., ``dress'', ``accessories'', ``digital''. Overall, we report the Accuracy@1 result $82.3\%$ of our approach for category prediction, which significantly increase the online model by over $9$ points. Specially, for the most confusable category, we increase the Precision@1 from 45.36\% to 55.93\% in ``others'', which shows the significant improvement by employing Virtual ID.

\begin{table*}[t]
\caption{Performance comparison for Virtual ID Category Network.}
\label{tb:CateResult}
\begin{center}
\resizebox{0.95\linewidth}{!}{%
\begin{tabular}{l||c|c|c|c|c|c|c|c|c|c|c|c|c|c||c}
\toprule
Method & shirt & dress & pants & bags & shoes & accessories & snacks &cosmetics& beverages & furniture & toys & underdress &digital & others & Accuracy \\
\hline
\hline
Online Model~\cite{zhang2018visual}&0.7604	&0.6573&	0.9480&	0.8415&	0.9574&	0.9308&	0.5906	&0.7831&	0.7127&	0.7739&	0.6523&	0.4507&	0.3568&	0.4536&	0.7288\\
\hline
Virtual ID branch &0.7986&	0.7466&	0.9440	&0.9456	&0.9712	&0.9590&	0.8148&	0.7895&	0.9473	&0.8926&	0.8839&	0.8247&	0.8279&	0.5057&	0.7648\\
\hline
Top Category branch & 0.7850 &	0.7539	& 0.9303&	0.9491&	0.9627&	0.9614	&0.7921&	0.7891&	0.9286&	0.8787&	0.8576&	0.7694&	0.8303&	0.4070&	0.7565\\
\hline
\hline
\textbf{Ours} & 0.8275	&0.7443&0.9418&	0.9001&	0.9575&	0.9578&	0.8081&	0.8548&	0.8624	&0.8168	&0.8287&	0.7762	& 0.7317&	0.5593	&\textbf{0.823}\\
\bottomrule
\end{tabular}}
\end{center}
\end{table*}

\textbf{Qualitative comparison.} To further illustrate the performance intuitively, we also visualize the predicted category from both online model and ours with corresponding ground truth labels in Figure~\ref{fig:searchRet}. From these cases, we notice that confusable leaf-categories are effectively identified by Virtual ID Category Network, due to that we consider more fine-grained partition of the categories rather than that online model considers only 14 top categories trained with human labeled data. With the regularization of the top category branch, hard samples are able be to learned to correct some misclassifications, such as objects in cell phone, unusual accessories, etc. Coarser and finer categories are considered complementarily to get better predicted result, which results in that the predictions are much closer to the users' intention.

\begin{figure}[t]
\begin{center}
\includegraphics[width=1.0\linewidth]{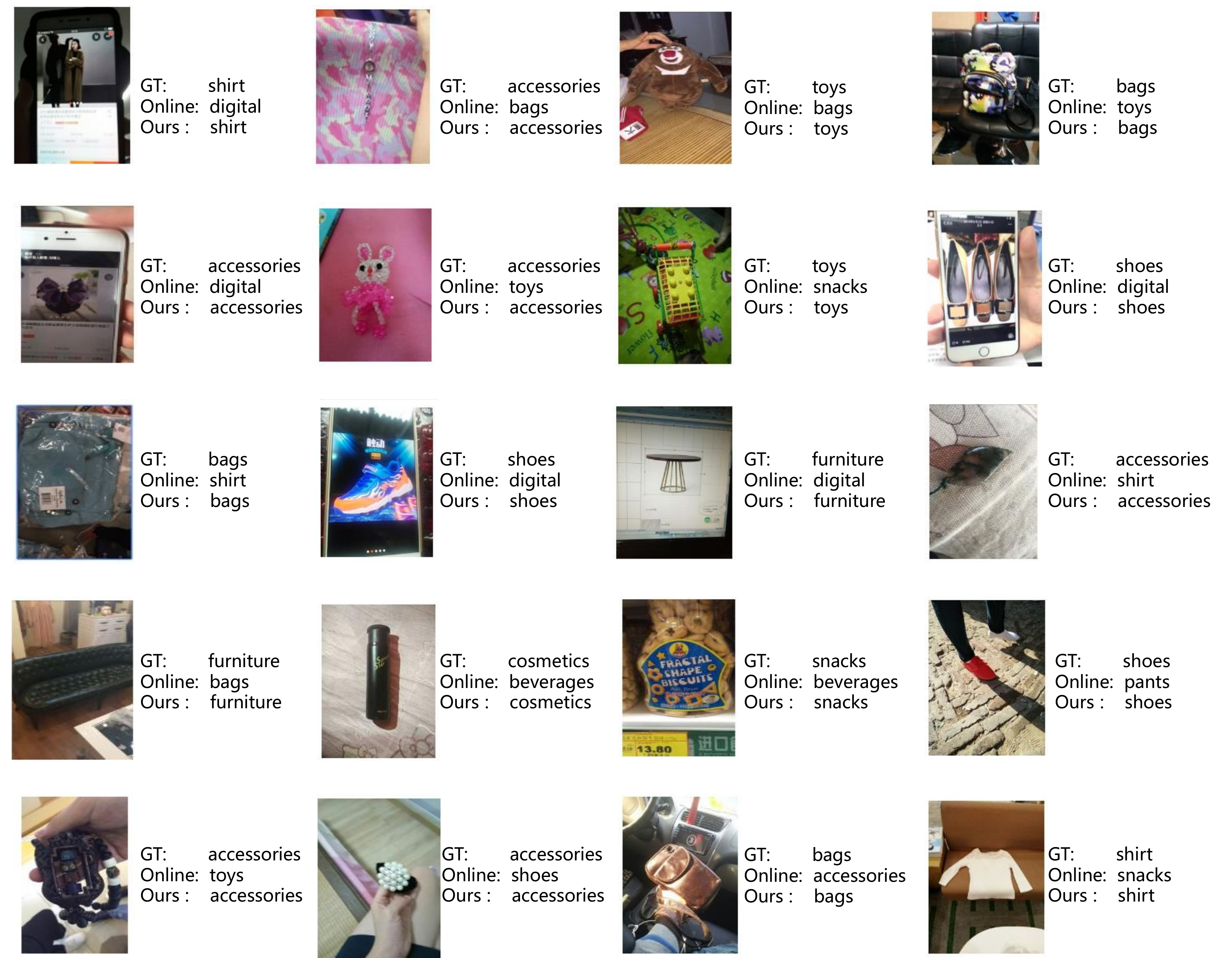}
\caption{Samples of category prediction results for real-shot images. }
\label{fig:objdet}
\end{center}
\end{figure}

\begin{figure}[t]
\begin{center}
\includegraphics[width=1.0\linewidth]{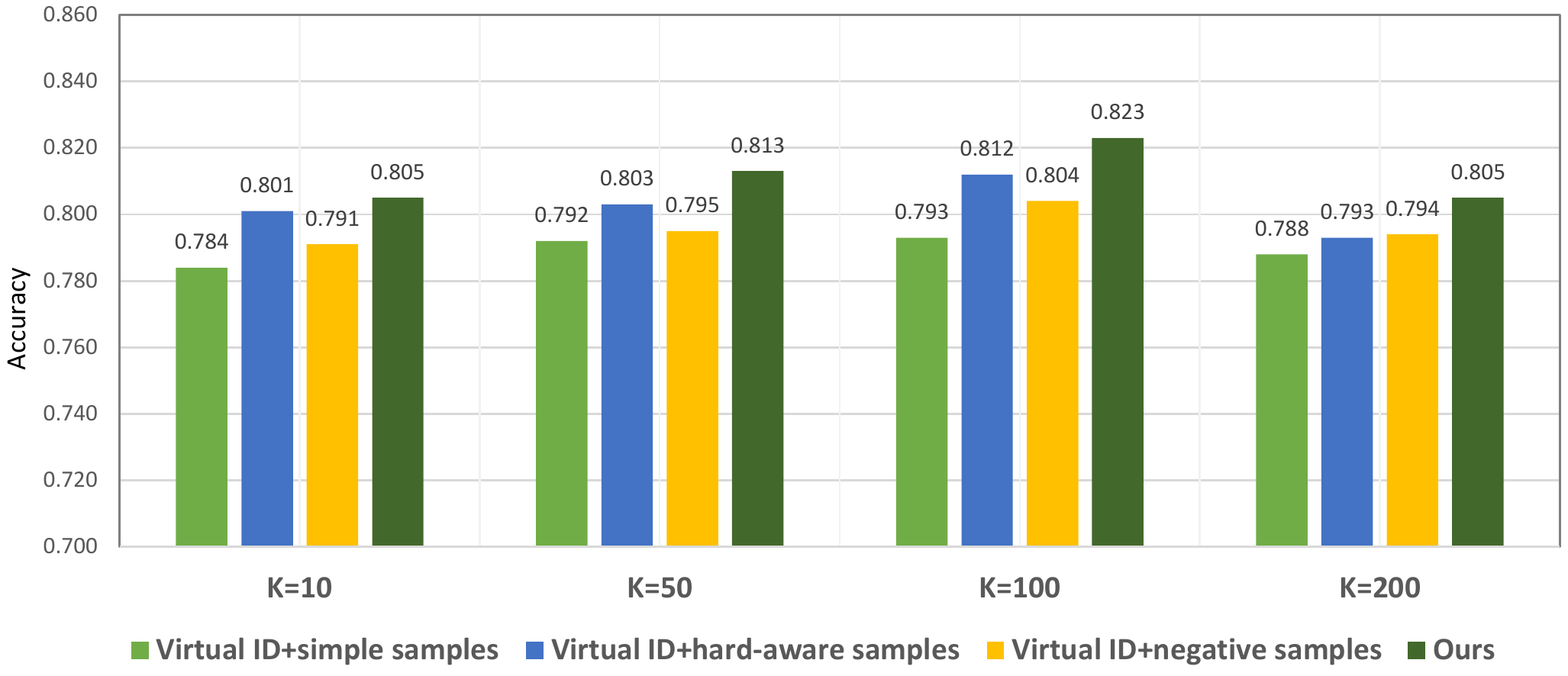}
\caption{Comparison among Virtual ID with different click samples in terms of category prediction accuracy.}
\label{fig:cat_param}
\end{center}
\vspace*{-3ex}
\end{figure}

\textbf{Effect of hyper-parameters.} As shown in Figure~\ref{fig:cat_param}, we report that results of our Virtual ID Networks by varying the number $K$ of Virtual ID. To predict the results, we can tell from the bars that ours achieves accuracy 80.5\%, 81.3\%, 82.3\% and 80.5\% for K=10,50,100,200, while K=100 achieves the best for the cluster configuration of Taobao leaf-category setting. Integrating with various click samples by tuning $\alpha$ and $\beta$ in Equation~\ref{eqn:loss_cat}, the comparison results present the progressive improvements from simple to hard and negative samples combined with Virtual ID, which indicates the value of switching-category data to regularize the Virtual ID branch and reduce generalization error.
\subsection{Evaluation of Feature Learning}
To evaluate the performance of Virtual ID Feature Network, we denote each real-shot image as the query to search for similar images in the item inventory. Following the same criterion to evaluate the learned feature~\cite{zhang2018visual}, we measure the relevance by Recall@K. As the metric to evaluate capability of feature, number of identical items returned among Top-K retrieved are counted. As the most revelent results, identical items clicked by the users' intention will produce the most possible conversions.

\textbf{Effect of Virtual ID branch.} Excluding the interference of category prediction, we validate the Feature Network by searching the image in the fixed category inventory. As the baselines, we compared our individual Virtual ID branch (trained with Virtual feature ID only) with start-of-the-art model-based results on real-shot images. The Virtual Feature ID is trained on 50M images with number K=0.5M of Virtual ID. The compared models are trained using PVLOG-triplets mining protocol proposed in~\cite{zhang2018visual}  with various backbone networks~\cite{krizhevsky2012imagenet,cvpr_SzegedyLJSRAEVR15,cvpr_HeZRS16}. Table~\ref{tb:compareVIDFeat} shows the performances of compared models, along with Identical Recall@K(K=1,4,20) based on the same 512-dim activations of FC layer. Instead of both branches, we note that Virtual ID branch only can compete against PVLOG-triplet data with larger model ResNet101, which shows significant improvements compared with the alternative backbones. 
\begin{table*}[t]
\caption{Performance comparison of Virtual ID Feature Network on real-shot image dataset.}
\label{tb:highrecallsetResult}
\begin{center}
\resizebox{1.0\linewidth}{!}{%
\begin{tabular}{l|l|l||c|c|c|c|c|c|c|c|c|c|c|c|c|c||c}
\toprule
Evaluation & Metric & Methods & shirt & dress & pants & bags & shoes & accessories & snacks &cosmetics& beverages & furniture & toys & underdress &digital & others & Average \\
\hline
\hline
\multicolumn{1}{l|}{\multirow{9}{*}{(A)Fixed category}}& \multirow{3}{*}{Recall@1}&Online\cite{zhang2018visual}& 65.80\%&  70.30\%&  68.40\%&  75.50\%&  63.80\%&  37.20\%&  58.40\%&  68.90\%&  52.10\%&  33.30\%&  70.30\%&  21.00\%&  36.90\%&  58.60\%&  61.38\%\\
\cline{3-18}
\multicolumn{1}{l|}{}&  &Ours&67.50\%&  73.60\%&  70.00\%&  76.90\%&  65.20\%&  39.00\%&  59.80\%&  70.00\%&  52.80\%&  36.30\%&  72.90\%&  22.00\%&  38.70\%&  60.20\%&  62.77\%\\
\cline{3-18}
\multicolumn{1}{l|}{}& & gain& {\color{red}+1.70\%}&{\color{red}+3.30\%}&{\color{red}+1.60\%}&{\color{red}+1.40\%}&{\color{red}+1.40\%}&{\color{red}   1.80\%}&{\color{red}+1.40\%}&{\color{red}+1.10\%}&{\color{red}+0.70\%}&{\color{red}+3.00\%}&{\color{red}+2.60\%}&{\color{red}+1.00\%}&{\color{red}+1.80\%}&{\color{red}+1.60\%}&{\color{red}+1.39\%}\\
\cline{2-18}
\multicolumn{1}{l|}{}&\multirow{3}{*}{Recall@4} &Online\cite{zhang2018visual}&79.10\%&  81.90\%&  83.60\%&  83.40\%&  78.90\%&  54.70\%&  63.10\%&  75.80\%&  62.20\%&  43.70\%&  81.70\%&  36.00\%&  53.20\%&  68.10\%& 72.47\%\\
\cline{3-18}
\multicolumn{1}{l|}{}& & Ours& 80.50\%&  85.70\%&  84.60\%&  84.10\%&  80.30\%&  56.50\%&  63.70\%&  76.60\%&  62.70\%&  45.90\%&  82.80\%&  40.00\%&  55.40\%&  68.30\%&  73.59\%\\
\cline{3-18}
\multicolumn{1}{l|}{}& &gain &{\color{red}+1.40\%}&{\color{red}+3.80\%}&{\color{red}+1.00\%}&{\color{red}+   0.70\%}&{\color{red}+1.40\%}&{\color{red}+1.80\%}&{\color{red}+0.60\%}&{\color{red}+0.80\%}&{\color{red}+   0.50\%}&{\color{red}+2.20\%}&{\color{red}+1.10\%}&{\color{red}+4.00\%}&{\color{red}+2.20\%}&{\color{red}+   0.20\%}&{\color{red}+1.12\%}\\
\cline{2-18}
\multicolumn{1}{l|}{}&\multirow{3}{*}{Recall@20} &Online\cite{zhang2018visual}&88.90\%&  90.80\%&  90.40\%&  88.70\%&  87.20\%&  69.20\%&  64.30\%&  79.80\%&  68.30\%&  53.30\%&  87.90\%&  48.00\%&  64.30\%&  73.10\%&  79.73\%\\
\cline{3-18}
\multicolumn{1}{l|}{}& & Ours&90.70\%&  91.70\%& 92.80\%&  89.80\%&  88.10\%&  73.70\%&  65.10\%&  81.10\%&  69.00\%&  54.10\%&  88.60\%&  50.00\%&  64.40\%&  73.30\%&  81.81\%\\
\cline{3-18}
\multicolumn{1}{l|}{}& &gain &{\color{red}+1.80\%}&{\color{red}+0.90\%}&{\color{red}+2.40\%}&{\color{red}+1.10\%}
&{\color{red}+0.90\%}&{\color{red}+4.50\%}&{\color{red}+1.80\%}&{\color{red}+1.30\%}&{\color{red}+0.70\%}
&{\color{red}+0.80\%}&{\color{red}+0.70\%}&{\color{red}+2.00\%}&{\color{red}+0.10\%}&{\color{red}+0.20\%}
&{\color{red}+2.08\%}\\
\hline
\hline
\multicolumn{1}{l|}{\multirow{9}{*}{(B)End to end offline}}& \multirow{3}{*}{Recall@1}&Online\cite{zhang2018visual}& 65.60\%&  69.30\%&  68.80\%&  72.70\%&  63.50\%&  36.70\%&  56.90\%&  70.30\%&  56.60\%&  30.70\%&  69.90\%&  18.00\%&  42.70\%&  49.00\%&  59.06\%\\
\cline{3-18}
\multicolumn{1}{l|}{}&  &Ours& 67.70\%&  72.60\%&  70.00\%&  73.90\%&  64.60\%&  38.90\%&  58.80\%&  71.90\%&  57.90\%&  35.60\%&  71.40\%&  20.00\%&  44.10\%&  52.50\%&  61.20\%\\
\cline{3-18}
\multicolumn{1}{l|}{}& & gain&{\color{red}+2.10\%}&{\color{red}+3.30\%}&{\color{red}+1.20\%}&{\color{red}+ 1.20\%}&{\color{red}+1.10\%}&{\color{red}+2.20\%}&{\color{red}+1.90\%}&{\color{red}+1.60\%}&{\color{red}+1.30\%}&{\color{red}+4.90\%}&{\color{red}+1.50\%}&{\color{red}+2.00\%}&{\color{red}+1.40\%}&{\color{red}+3.50\%}&{\color{red}+2.14\%}\\
\cline{2-18}
\multicolumn{1}{l|}{}&\multirow{3}{*}{Recall@4} &Online\cite{zhang2018visual}&79.40\%&  81.30\%&  81.90\%&  80.20\%&  79.70\%&  54.20\%&  61.00\%&  75.90\%&  65.90\%&  39.40\%&  80.80\%&  32.30\%&  58.10\%&  58.00\%&  69.07\%\\
\cline{3-18}
\multicolumn{1}{l|}{}& & Ours& 80.70\%&  85.10\%&  84.50\%&  81.20\%&  79.90\%&  57.20\%&  62.70\%&  78.70\%&  68.60\%&  44.40\%&  81.30\%&  40.00\%&  58.60\%&  59.30\%&  71.83\%\\
\cline{3-18}
\multicolumn{1}{l|}{}& &gain &{\color{red}+1.30\%}&{\color{red}+3.80\%}&{\color{red}+2.60\%}&{\color{red}+1.00\%}&{\color{red}+0.20\%}&
{\color{red}+3.00\%}&{\color{red}+1.70\%}&{\color{red}+2.80\%}&{\color{red}+2.70\%}&{\color{red}+5.00\%}
&{\color{red}+0.50\%}&{\color{red}+7.70\%}&{\color{red}+0.50\%}&{\color{red}+1.30\%}&{\color{red}+2.76\%}\\
\cline{2-18}
\multicolumn{1}{l|}{}&\multirow{3}{*}{Recall@20} &Online\cite{zhang2018visual}&88.70\%&  90.20\%&  89.80\%&  86.50\%&  86.90\%&  67.70\%&  62.00\%&  79.60\%&  71.60\%&  48.00\%&  87.70\%&  51.00\%&  62.50\%&  62.30\%&  76.74\%\\
\cline{3-18}
\multicolumn{1}{l|}{}& & Ours& 91.40\%&  92.20\%&  93.50\%&  88.80\%&  87.40\%&  74.20\%&  64.10\%&  83.70\%&  74.70\%&  54.10\%&  88.60\%&  53.00\%&  63.10\%&  63.20\%&  79.14\%\\
\cline{3-18}
\multicolumn{1}{l|}{}& &gain &{\color{red}+ 2.70\%}&{\color{red}+2.00\%}&{\color{red}+3.70\%}&{\color{red}+   2.30\%}&{\color{red}+0.50\%}&{\color{red}+6.50\%}&{\color{red}+2.10\%}&{\color{red}+4.10\%}&{\color{red}+3.10\%}&{\color{red}+6.10\%}&{\color{red}+0.90\%}&{\color{red}+2.00\%}&{\color{red}+0.60\%}&{\color{red}+0.90\%}&{\color{red}+2.40\%}\\
\hline
\hline
\multicolumn{1}{l|}{\multirow{2}{*}{(C)Online A/B test}}&CTR&Ours &+5.54\%&+4.35\%&+6.25\%&+4.53\%&+7.15\%&	+6.54\%	&+5.49\%&+6.37\%	&+8.26\%&+8.43\%	&+6.49\%&+5.33\%	&+5.17\%	&+7.63\%	&\textbf{+5.86\%}\\
\cline{2-18}
\multicolumn{1}{l|}{}&GMV &Ours &+6.23\%&+5.34\%&+8.23\%&+5.27\%&+6.87\%&	+9.54\%	&+4.47\%&+7.39\%	&+8.54\%&+9.21\%&+6.19\%&+5.32\%	&+7.19\%	&+8.37\%	&\textbf{+7.23\%}\\
\bottomrule
\end{tabular}}
\end{center}
\end{table*}

Also, we report the overall results of our visual feature on all categories in Table~\ref{tb:highrecallsetResult}(A). For all experiments, we search for 20 similar images within each predicted category for Recall@K (K=1,4,20). Identical Recall of our approach improves as $K$ increases, which clearly shows that our approach introduces more relevant images into the top search results. Compared to online model trained with PVLOG-triplets, we are able to achieve better performance when retrieving with our learned feature.
\begin{table}[t]
\caption{Comparisons of different visual features on real-shot images.}
\label{tb:compareVIDFeat}
\begin{center}
\resizebox{0.8\linewidth}{!}{%
\begin{tabular}{l|l|c|c|c}
\toprule
\multicolumn{2}{l|}{Model}& Recall@1 & Recall@4 & Recall@20 \\
\hline
\hline
\multirow{4}{*}{PVLOG-triplets~\cite{zhang2018visual} }&AlexNet~\cite{krizhevsky2012imagenet}& 0.346 &0.423&0.553\\
\cline{2-5}
&GoogLeNet~\cite{cvpr_SzegedyLJSRAEVR15}&0.465 &0.564 &0.629 \\
\cline{2-5}
&ResNet50~\cite{cvpr_HeZRS16} & 0.585 &0.685 &0.762\\
\cline{2-5}
&ResNet101~\cite{cvpr_HeZRS16} & 0.613 &0.718 &0.784\\
\hline
\hline
\multicolumn{2}{l|}{Virtual ID data + ResNet50(Ours)} & \textbf{0.615}& \textbf{0.715}& \textbf{0.798}  \\
\bottomrule
\end{tabular}}
\end{center}
\end{table}

\begin{figure}[t]
\begin{center}
\includegraphics[width=1.0\linewidth]{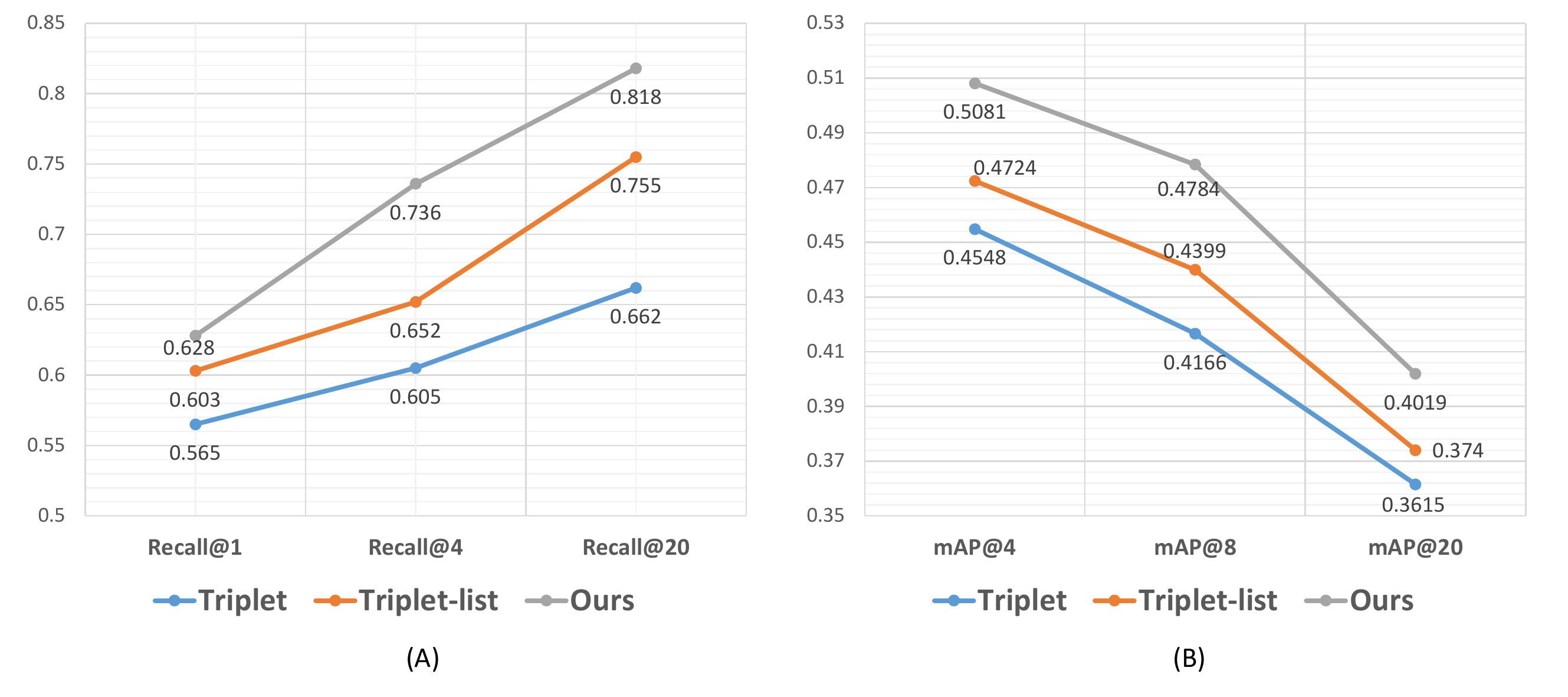}
\caption{Comparison among features with triplet and list samples on A) Recall (B) mAP. }
\label{fig:exp-2}
\end{center}
\vspace*{-3ex}
\end{figure}

\textbf{Effect of ranking branch.} As illustrated in Section~\ref{sec:joinmodel}, we found the ranking loss preserves similarity relations within the retrieval list, so we jointly train the deep networks by sampling from list images for valid triplets and list samples without further annotations. To validate the superiority of the triplet-list samples, we compared the models with alternative models using these separated samples that are trained without Virtual ID data. As shown in Figure~\ref{fig:exp-2}(A), we outperform list-based variants and increase the Identical Recall@1 by over 4 and 2 percentage points when compared with triplets and triplet-list. Clearly that triplet-list encodes more information from clicked images for discriminating identical items. Besides, in terms of mean Average Precision(mAP) metric in Figure~\ref{fig:exp-2}(B), we surpass the feature with triplets by 4\% mAP@1 and triplet-list by 3\% mAP@1, indicating more relevant images are retrieved.

\begin{figure*}[t]
\begin{center}
\includegraphics[width=1.0\linewidth]{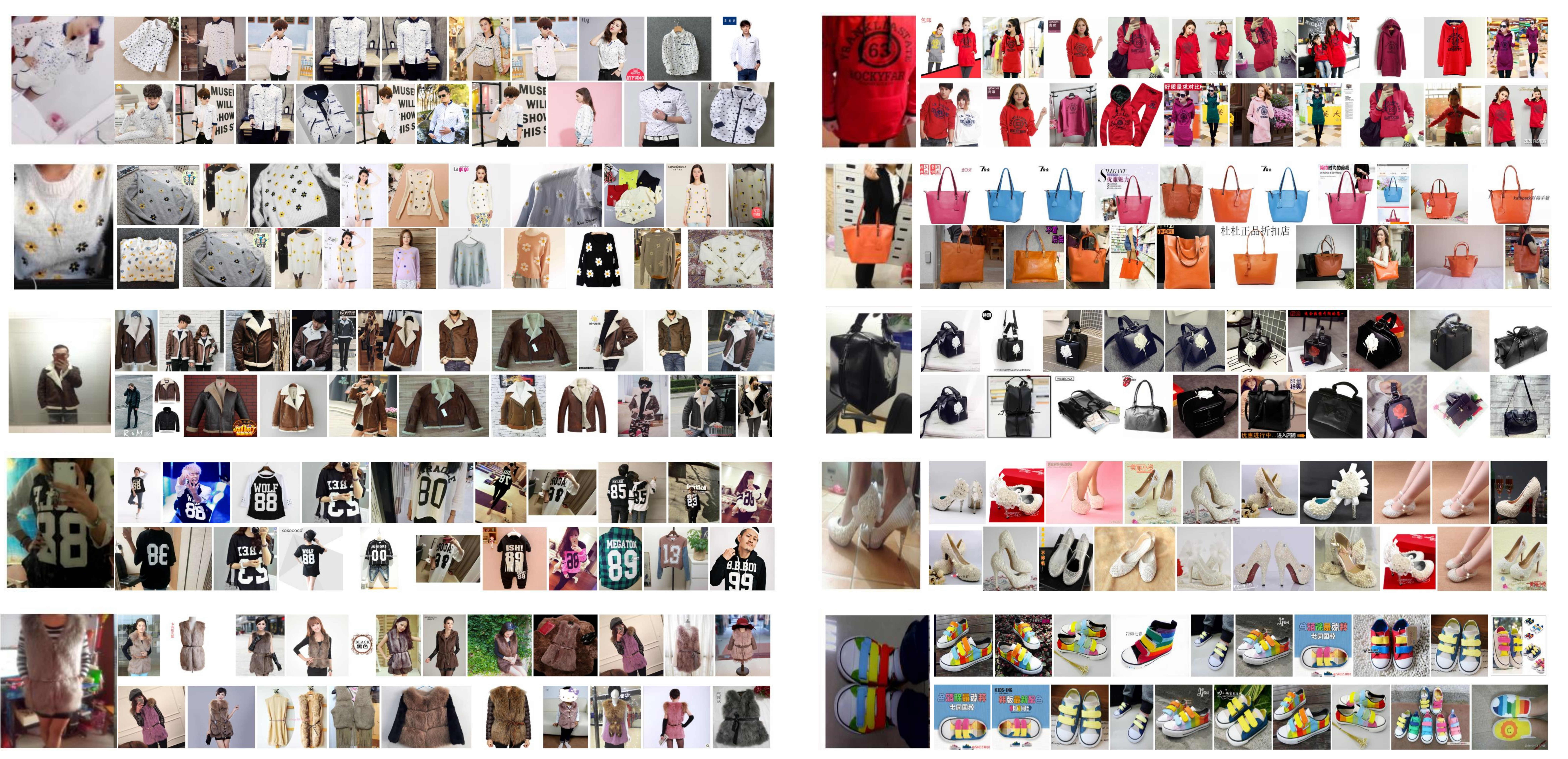}
\caption{Qualitative comparsion of search results. Real-shot query images are followed by top 10 ranked images from active listings. Top row: retrieved list from Virtual ID Feature Network, bottom row: the list from online model.}
\label{fig:searchRet}
\end{center}
\end{figure*}

\textbf{Qualitative comparison.} In Figure~\ref{fig:tsne}, we further validate the fine-grained discriminating capacity of our learning embedding intuitively. We employ tSNE~\cite{van2008visualizing} to  illustrate the 512-dim semantic feature space for 6 leaf-categories of ``skirt'' including underclothes, wind coat, hoodie, jacket, feather dress and fur coat. From the distribution, we claim  that our feature preserves semantic information even in the confusable neighborhood qualitatively. In Figure~\ref{fig:searchRet}, we visualize the retrieval results for real-shot query images, which presents satisfying returned list on identical items. As we can see,  the returned lists from ours achieve more identical items to the top positions and perform more consistently than online model in terms of proper order and similar ratio. It contributes to that our approach exerts more relationship information from click data,  and more discriminative to recall semantically similar items both in local details and general styles.

\subsection{End to End Search Relevance}
Coupling with category prediction, we state Recall metric of end to end structure compared with online model, which is utilized on 3 billion images in Table~\ref{tb:highrecallsetResult}(B). Compared with online model, the overall result of end to end evaluation is increased by over 2\% in terms of Recall@1. Recall of different categories is more diverse, while the least ``shoes'' still outperforms online by 1.1\%, due to it is easier to find the identical in this category. Besides, we present the results of online evaluation in Pailitao with a standard A/B testing configuration. We concern most on two important commercial metrics, i.e. Click Through Rate(CTR) and Gross Merchandise Volume(GMV). Table~\ref{tb:highrecallsetResult}(C) reports the results of improvements on CTR, GMV when we apply Virtual ID Networks on the operational visual search system. The results show that CTR can increase 5.86\% and GMV can increase 8.43\% in average, which provides users with more acceptable and enjoyable visual search experience. Table~\ref{tb:highrecallsetResult}(C) also presents a more detailed result of the increase on each category, achieving a maximum 8.43\% CTR increase in ``furniture'' and 9.54\% GMV engagement in ``accessories''.



\section{Conclusions}
This paper propose a click-data driven approach for visual search relevance at Alibaba. Effective co-click embeddings that are learned as Virtual ID help to supervise classification for category prediction and feature learning, which explores underlying relationship among images. By exploiting various types of click data, first-clicks and switch-clicks facilitate to regularize Virtual ID Category Network. Besides, Virtual ID Feature Network is designed in a joint classification and ranking manner by integrating triplet-list constraint. Benefiting from richness of these click data, visual search relevance can be more effective to reflect users' interests. We further provide some practical lessons of learning and deploying the deep networks in the visual search system.

Extensive experiments on real-shot images are provided to illustrate the the competitive performance of Virtual ID Networks as well as generality and transferability of the click-data based approach. Detailed discussion of some case studies are also provided to show the insight of how each component affects. Virtual ID discovery has already been deployed successfully on the online service in visual search system at Alibaba. Online A/B testing results show that our models increase users' preference better and improve shopping conversions. In our future work, interactive search and image tagging will be leveraged in Pailitao to advocate visual search relevance.